\documentclass[conference]{IEEEtran}
\IEEEoverridecommandlockouts
\usepackage{cite} 
\usepackage{amsmath,amssymb,amsfonts}
\usepackage{algorithmic}
\usepackage{graphicx}
\usepackage{textcomp}
\usepackage{xcolor}
\usepackage{dsfont}
\usepackage{wrapfig}
\usepackage{xcolor}
\usepackage{colortbl}
\usepackage{booktabs}
\usepackage{soul}
\usepackage{hyperref}
\usepackage{cleveref}
\usepackage{enumitem}
\usepackage{balance}

\hypersetup{
     colorlinks   = true,
     citecolor    = black,
     linkcolor    = black,
     urlcolor = black,
}

\usepackage[linesnumbered]{algorithm2e}
\SetKwComment{Comment}{// }{}
\RestyleAlgo{ruled}

\let\orgautoref\autoref

\renewcommand{\autoref}[1]
{%
\def\equationautorefname{Eq.}%
\def\figureautorefname{Fig.}%
\def\subfigureautorefname{Fig.}%
\def\sectionautorefname{Sec.}%
\def\subsectionautorefname{Sec.}%
\def\algorithmautorefname{Alg.}%
\orgautoref{#1}%
}

\newcommand{\snnote}[1]{\textcolor{orange}{}}
\newcommand{\ndedit}[1]{\textcolor{black}{#1}}

\begin{document}

\title{Contrastive Learning from Exploratory Actions: Leveraging Natural Interactions\\for Preference Elicitation}

\author{\IEEEauthorblockN{Nathaniel Dennler}
\IEEEauthorblockA{
\textit{University of Southern California} \\
Los Angeles, California, USA\\
\texttt{dennler@usc.edu}\vspace{-2em}}
\and
\IEEEauthorblockN{Stefanos Nikolaidis}
\IEEEauthorblockA{
\textit{University of Southern California} \\
Los Angeles, California, USA\\
\texttt{nikolaid@usc.edu}\vspace{-2em}}
\and
\IEEEauthorblockN{Maja Matari\'c}
\IEEEauthorblockA{
\textit{University of Southern California} \\
Los Angeles, California, USA\\
\texttt{mataric@usc.edu}\vspace{-2em}}
}

\maketitle

\begin{abstract}
People have a variety of preferences for how robots behave.
To understand and reason about these preferences, robots aim to learn a reward function that describes how aligned robot behaviors are with a user's preferences. 
Good representations of a robot's behavior can significantly reduce the time and effort required for a user to teach the robot their preferences. 
Specifying these representations---what ``features" of the robot's behavior matter to users---remains a difficult problem;
Features learned from raw data lack semantic meaning and features learned from user data require users to engage in tedious labeling processes.
Our key insight is that users tasked with customizing a robot are intrinsically motivated to produce labels through \textit{exploratory search}; they explore behaviors that they find interesting and ignore behaviors that are irrelevant.
To harness this novel data source of \textit{exploratory actions}, we propose \textit{contrastive learning from exploratory actions} (CLEA) to learn trajectory features that are aligned with features that users care about.
We learned CLEA features from exploratory actions users performed in an open-ended signal design activity (\textit{N}=25) with a Kuri robot, and evaluated CLEA features through a second user study with a different set of users (\textit{N}=42).
CLEA features outperformed self-supervised features when eliciting user preferences over four metrics: completeness, simplicity, minimality, and explainability. 

\end{abstract}



\section{Introduction}

People have a variety of preferences for how robots should behave based on many contextual factors, but those contextual factors are often unknown to the designers of robotic systems before a robot is deployed. 
Consider a wheeled robot that helps users find misplaced items in their home. 
One user may be a long-time dog owner and thus interpret this interaction as similar to playing fetch. 
That user might expect the behavioral aspects of the robot to be dog-like.
For example, the robot may move erratically as if following a scent, bark when it has found an item, and emote to portray happiness having completed its command.
Another user, in contrast, may be more familiar with smart devices and expect the interaction to be purely functional.
That user might instead expect the robot to move and scan the room methodically, chime when it finds an item, and immediately return the item to the requester.

\begin{figure}
         \centering
         \includegraphics[width=\linewidth]{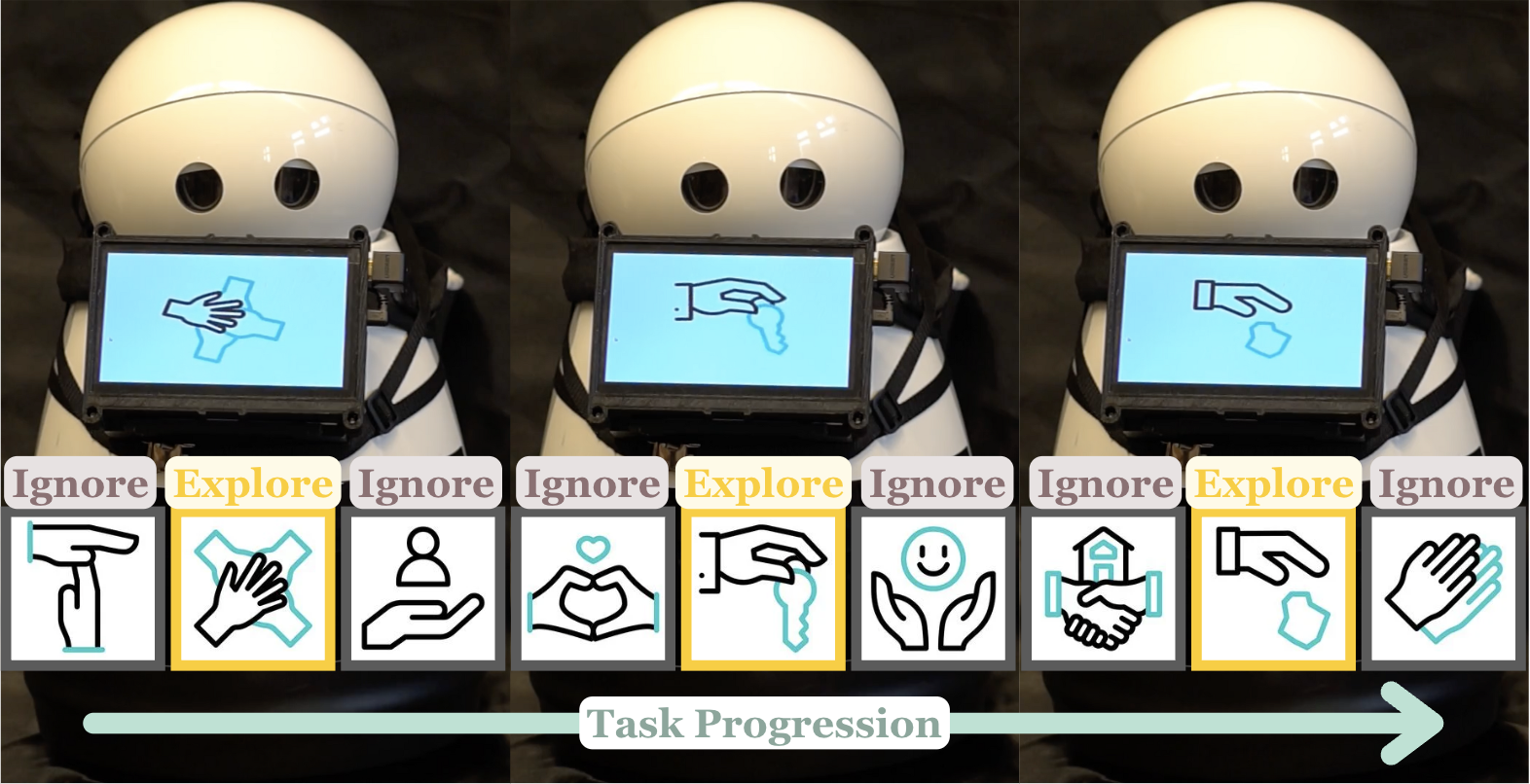}
         \caption{\textbf{Example exploratory search process}. Users engaging in \textit{exploratory search} test out different robot behaviors to learn what the robot is capable of and what they prefer the robot to do.}
         \label{fig:exploratory_example}
         \vspace{-1em}
\end{figure}

Deploying a robot that behaves in only one way cannot satisfy both of these users. 
Thus, users must be able to \textit{customize} robot behaviors to align with their preferences.
Several works view the problem of aligning the robot with the user's preferences as modeling the user's internal reward function, which can be addressed with inverse reinforcement learning \cite{ng2000algorithms,abbeel2004apprenticeship}. In this context, the reward function takes in numerical ``features" of the robot's behavior, e.g., a score of how dog-like or machine-like the behavior is, and output a single value that corresponds to how good that behavior is for the user.
How these features are defined heavily influences how effectively a robot can adapt to a specific user. 
Features can be learned directly from the robot behaviors through self-supervised techniques like autoencoders (AEs) and variational autoencoders (VAEs).
While self-supervised methods result in features that are physically representative of the robot's behaviors, they may not align with the features people actually care about.
The most effective way to learn user-aligned features is by leveraging user-generated data \cite{bobu2024aligning}.
However, collecting such data typically requires a user to engage in a data-labelling process known as a \textit{proxy task} before the user can engage in the actual task of customizing the robot \cite{bobu2023sirl,lee2021pebble,yang2021representation}.

\begin{figure*}
         \centering
         \includegraphics[width=.9\linewidth]{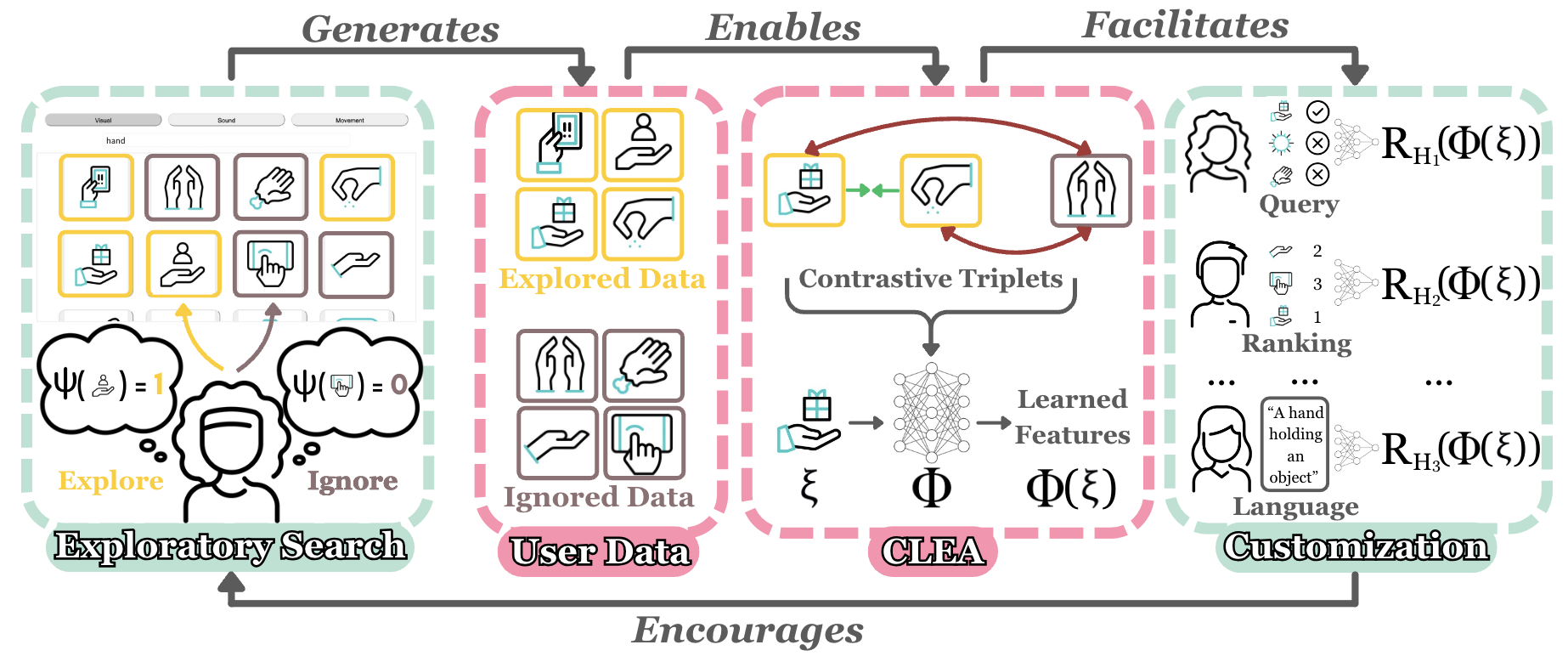}
         \caption{\textbf{CLEA: Contrastive Learning from Exploratory Actions}. Users engage in exploratory search to select their preferred robot behaviors. We automatically generate data from their exploratory actions to learn features that facilitate future interactive learning processes. Our contributions are highlighted in pink, and the enabling work that CLEA supports in highlighted in green.}
         \label{fig:framework}
         \vspace{-1em}
\end{figure*}

Our goal in this work is to learn features for robot behaviors that are aligned with user preferences, but do not require users to engage in unrelated proxy tasks.
To accomplish this, we identify a new form of user-collected data that is generated during the robot customization process.
We collected this data by recruiting users to customize behaviors for a Mayfield Kuri robot that helped them locate items around a room.
Participants used the RoSiD interface~\cite{dennler2023rosid} to design state-expressive signals, which allowed them to search through thousands of possible robot behaviors.
Participants automatically performed \textit{exploratory actions} by selecting behaviors that appeared interesting to them while ignoring those they thought were irrelevant, as illustrated in \autoref{fig:exploratory_example}.

Our key insight is that we can use these exploratory actions to learn features for robot behaviors that are both aligned with the features that users care about and do not require users to complete proxy tasks before customizing the robot. 
We view users performing exploratory actions as engaging in an intuitive reasoning process, and model this using a contrastive loss to train feature-generating networks.
We call this framework \textit{contrastive learning from exploratory actions} (CLEA) and provide an overview in \autoref{fig:framework}.

We show that CLEA learns features that are more effective for eliciting a user's preferences than the state-of-the-art self-supervised learning techniques, offering a scalable and user-friendly approach to personalizing robot behaviors.
\ndedit{In \autoref{sec:data_collection}, we collected a training set of exploratory actions from 25 participants specifying their signaling preferences to learn CLEA features.
In \autoref{sec:experiments}, we evaluated the generalizability of those features with a testing set from 42 na\"ive participants.} 
We found that CLEA features outperformed self-supervised features for robot signaling along four criteria \cite{bobu2024aligning}: they contained information relevant to elicit user preferences, required fewer user interactions to elicit preferences, captured information in fewer dimensions, and demonstrated properties that indicate compatibility with explanatory methods.


\section{Related Work}

\noindent\textbf{Eliciting User Preferences through Interaction.} 
Users must be able to communicate their preferences to the robot so the robot can learn these preferences. 
Previous works in preference learning identified several interactions that allow users to specify their preferences for robot behaviors, including behavior comparisons \cite{biyik2019asking, sadigh2017active, lu2022preference}, behavior rankings \cite{brown2019extrapolating, myers2022learning,chen2021learning}, binary rewards \cite{celemin2019interactive}, corrections \cite{bajcsy2018learning,li2021learning,cui2018active,losey2018including}, natural language \cite{tabrez2021asking,raman2022planning,liu2023grounding,wu2023tidybot}, facial expression \cite{cui2021empathic,stiber2023using}, and demonstrations \cite{nikolaidis2013human,nikolaidis2015efficient,ziebart2008maximum,cao2021learning,hughes2020inferring}. 
Those interactions require different skills and provide varying levels of information on the user's true preferences \cite{fitzgerald2022inquire,biyik2022learning,cui2021understanding}.
Additionally, all of them assume that there is a numerical representation of robot behavior that encapsulates the features that users care about. Thus, having meaningful numerical features is necessary to allow users of different skill levels to teach robots in a variety of ways.

\noindent\textbf{Learning Representations for Eliciting Preferences.} 
There are three popular approaches for representing features of robot behaviors: hand-crafted features, features learned from modeling robot behaviors, and features learned from user interactions.
Hand-crafted features are based on an engineer's intuitions of what is meaningful for users~\cite{bajcsy2018learning,hadfield2017inverse,ng2000algorithms,huang2024modeling}. Such features can speed up the preference learning process because they are meaningful to users, but they can also be difficult to design and can lead to incomplete feature spaces that limit the range of preferences that can be captured~\cite{bobu2024aligning}.

In contrast, features learned from modeling the robot's behaviors require less engineering effort, but are not meaningful. Such techniques learn features with little human input through self-supervised learning \cite{brown2020safe,laskin2020curl,ha2018recurrent} or weakly-supervised learning \cite{wang2024rl,chamzas2020state,brehmer2022weakly}. While these algorithms result in features that describe the underlying behaviors well and do not require extensive data collection from users, the resulting feature spaces are not semantically meaningful to users~\cite{bobu2023sirl}.  

Learning feature spaces from human input can result in feature spaces that are both complete and meaningful. 
A user may manually select features \cite{bullard2018human,cakmak2012designing}, physically move compliant robots to demonstrate behaviors \cite{bobu2021feature}, provide demonstrations for use in multi-task learning \cite{nishi2020fine,yamada2022task} and meta-learning frameworks\cite{schrum2022mind,schrum2023reciprocal}, or answer trajectory similarity queries \cite{bobu2023sirl}. 
These methods focus on developing \textit{proxy tasks} to learn features that are aligned with user preferences~\cite{bobu2024aligning}, however these tasks require conscious effort from the user.  

We emphasize that such proxy tasks are not necessarily aligned with the users' goal of customizing a robot's behaviors, and thus users may be unmotivated to perform the tasks \cite{thompson1998extending,lee2015relating}.
\ndedit{In contrast, \textit{exploratory search} provides an interaction that allows the user to achieve their primary goal of robot behavior customization. This work identifies that \textit{exploratory search} additionally generates data that can be used to learn robot feature spaces in place of proxy-task data.}

\noindent\textbf{Exploratory Search.} Work in human-computer interaction (HCI) distinguishes two interactions with databases: \textit{information retrieval} and \textit{exploratory search} \cite{marchionini2006exploratory}. \textit{Information retrieval} \cite{sanderson2012history, singhal2001modern} refers to an interaction with a data system wherein the user knows exactly what they need to find---the user's reward is known.  In \textit{exploratory search}, the exact goal is unknown ahead of time because the user is unfamiliar with the search topic and how the goal can be achieved \cite{marchionini2006exploratory}. While many previous works implicitly assume that users know what the robot is capable of doing and present preference learning as an information retrieval problem \cite{ng2000algorithms,singhal2001modern}, recent work has identified that reformulating robot learning as an exploratory search interaction is useful for novice robot users \cite{dennler2023rosid}.

Exploratory search interfaces encourage users to generate more search data by allowing them to inspect, save, and filter items in large databases \cite{allen2021engage,chang2019searchlens,rahdari2020grapevine}.
By aggregating and scaling these search data across millions of users, HCI researchers can learn fine-grained profiles of user behaviors \cite{he2016beyond,yang2022click}.

In this work, we examine the effectiveness of exploratory actions for learning features of robot behaviors that users care about. We frame an exploratory action as an intuitive reasoning process where a user quickly evaluates if a robot behavior is somewhat aligned with their preferences. If it is, they select that behavior to perform a more in-depth evaluation on the physical robot. These perceptual processes are often modeled with triplet losses both to capture how people make intuitive decisions and to aggregate individual differences across user populations \cite{bobu2023sirl,demiralp2014learning,hadsell2006dimensionality,hoffer2015deep,radlinski2005query}. We use this insight from prior work to learn features of robot behaviors that people care about from the novel data source of exploratory actions. 


\section{Learning Features from Exploratory Actions}\label{sec:CLEA}

In this section, we formalize our approach that leverages exploratory actions to learn features of robot behaviors.

\subsection{Preliminaries} 
We consider robot behaviors as trajectories in a fully-observed deterministic dynamical system.
We denote a behavior as $\xi \in \Xi$, which represents a series of states and actions: $\xi = (s_0,a_0,s_1,a_1,...,s_T,a_T)$. 
\ndedit{These states and actions are abstractly defined; they can be videos (behaviors in image-space), audio (behaviors in frequency-space), or movements (behaviors in joint-space).}
We assume that all behaviors $\xi \in \Xi$ accomplish the task without resulting in errors, allowing users to specify based on user preferences rather than the behavior's ability to achieve a goal~\cite{nikolaidis2015efficient,nemlekar2023transfer,bobu2023sirl}. 
While generating $\Xi$ is not the focus of this work, it can be completed through several techniques, such as collecting demonstrations \cite{nikolaidis2015efficient}, performing quality diversity optimization \cite{tjanaka2022approximating}, and diversely combining motion primitives \cite{wang2018active}.

We model a user's preference as a reward function over robot behaviors that maps the space of behaviors to a real value: $R_H : \Xi \mapsto \mathbb{R}$. 
The user's reward function is not directly observable, but can be inferred through interaction. Our goal is to learn a reward function from user interactions, $R_H$, that maximizes the likelihood of the user performing the observed interactions.
Higher values of $R_H$ for a particular behavior implies that the behavior is more preferred by the user.

Because the state space of robot behaviors can be very large \cite{robinson2023robotic,arulkumaran2017deep}, directly learning $R_H$ from state-action sequences is intractable. 
To make reward learning tractable, several works \cite{bobu2024aligning,ng2000algorithms,abbeel2004apprenticeship} assume that there exists a function $\Phi$ that maps from the state-action space to a lower dimensional \textit{feature space}--a real vector of dimension $d$: $\Phi : \Xi \mapsto \mathbb{R}^d$. 
This assumption allows us to learn $R_H(\Phi(\xi))$ from fewer user interactions.

\subsection{Contrastive Learning from Exploratory Actions} 

To learn a feature function, $\Phi$, we leverage interaction data that we collected through the robot customization process.
Users naturally engaged in \textit{exploratory search} when they were presented with many robot behaviors they could choose from to customize the robot.  

We formalize exploratory search as presenting a dataset of behaviors to the user: $\mathcal{D}_i = \{\xi_0, \xi_1,...,\xi_N\}$ where each $\xi_i$ is sampled from the full database of behaviors $\Xi$. In our case, $\xi_i$ is a video, a sound, or a head movement, but this definition extends to other behaviors such as robot gaits, or robot arm movements.
This dataset can be generated using various methods, including keyword search \cite{chapman2020dataset}, collaborative filtering \cite{koren2021advances}, and faceted search \cite{yee2003faceted}. 
Users can view brief summaries of each behavior in the dataset to determine if the behavior is relevant.

We mathematically model the user's internal reasoning process when making an exploratory action with the function $\psi : \mathcal{D} \mapsto \{0,1\}$.
If the user performs an exploratory action on a behavior $\xi_j$ from the dataset $\mathcal{D}_i$, then $\psi(\xi_j) = 1$.
If the user does not perform an exploratory action on a behavior $\xi_k$ from the dataset $\mathcal{D}_i$, then $\psi(\xi_k) = 0$.
We use this definition of an exploratory action to partition $\mathcal{D}_i$ into two sets:

\vspace{-1em}
\begin{equation}
    \mathcal{D}_i^{ex.} := \{ \xi \in \mathcal{D}_i | \psi(\xi) = 1\};  \mathcal{D}_i^{ig.} := \{ \xi \in \mathcal{D}_i | \psi(\xi) = 0\} 
\end{equation}

For example, if a user is initially presented with $\mathcal{D}_0 = \{\xi_A, \xi_B,\xi_C, \xi_D\}$, and they choose $\xi_B$ and $\xi_D$ to execute on the robot, the explored dataset is $\mathcal{D}_0^{ex.} = \{\xi_B, \xi_D\}$ and the ignored dataset is $\mathcal{D}_0^{ig.} = \{\xi_A, \xi_C\}$. 
In our data collection study, $|\mathcal{D}_i| \approx 100$ to allow users to meaningfully search through behaviors. 

A common way to model and aggregate diverse internal reasoning processes, such as $\psi$, across a population of users is to use a triplet loss \cite{bobu2023sirl,demiralp2014learning,hadsell2006dimensionality,hoffer2015deep,radlinski2005query}.
We adopt this loss function and generate triplets of behaviors from on the explored and ignored subsets. The triplets are formed by sampling two behaviors at random from one subset and one behavior from the other subset: $(\xi^{\mathcal{D}_i^{ex.}}_1, \xi^{\mathcal{D}_i^{ex.}}_2, \xi^{\mathcal{D}_i^{ig.}}_1)$ or conversely $(\xi^{\mathcal{D}_i^{ig.}}_1, \xi^{\mathcal{D}_i^{ig.}}_2, \xi^{\mathcal{D}_i^{ex.}}_1)$.
The triplet loss encourages features from the same subset to be more similar to each other than features from opposite subsets, according to any metric function. We select the Euclidean distance, $d(\xi_i, \xi_j) = ||\Phi(\xi_i) - \Phi(\xi_j)||_2^2$, as our metric because other works found that Euclidean distances are an appropriate metric for modeling perceptual processes \cite{bobu2023sirl,demiralp2014learning}:

\vspace{-1em}
\begin{equation}
    \mathcal{L}_{trip.}(\xi_A, \xi_P, \xi_N) = \max\left[d(\xi_A,\xi_P) - d(\xi_A,\xi_N)+ \alpha, 0\right]
\end{equation}

We refer to $\xi_A$ as the anchor example, $\xi_P$ as the positive example, $\xi_N$ as the negative example, and $\alpha \geq 0$ as the margin of separation between positive and negative examples. In our case, the anchor and positive example are interchangeable as they are both from the same unordered set, so we formulate the triplet loss to be symmetric:

\vspace{-1em}
\begin{equation}
    \mathcal{L}_{sym.}(\Phi) = \mathcal{L}_{trip.}(\xi_A, \xi_P, \xi_N) + \mathcal{L}_{trip.}(\xi_P, \xi_A, \xi_N)
\end{equation}

We formulate the CLEA loss as the sum of this symmetric triplet loss across all of the datasets presented to all the users in the signal design study:

\vspace{-1em}
\begin{equation}\label{eq:training_objective}
    \mathcal{L}_{CLEA}(\Phi) = \sum_{i=0}^{|\mathcal{D}_{pop.}|} \sum_{(\xi_A, \xi_P, \xi_N) \sim \mathcal{D}_i} \mathcal{L}_{sym.}(\xi_A, \xi_P, \xi_N)
\end{equation}

where $\mathcal{D}_{pop}$ represents the set of all datasets presented to the population of users that performed exploratory actions. 
We learn features that minimize this loss to create a feature space for robot behaviors that is consistent with the variations in the population's preferences.

\subsection{Learning Preferences from Rankings.} 
To evaluate CLEA on a new population and task, we used behavior rankings, as in previous works \cite{myers2022learning}.
We presented each user with a set of behaviors to rank, referred to as a query, $Q = \{\xi_0, \xi_1,...\xi_N\}$. 
The user then ordered these options from their least favorite behavior to their most favorite behavior by creating a mapping $\sigma : \{0,1,...,N\} \mapsto \{0,1,...,N\}$ such that $\sigma(Q) := \xi_{\sigma(0)} \prec \xi_{\sigma(1)} \prec ... \prec \xi_{\sigma(N)}$. The notation $\xi_i \prec \xi_j$ denotes that behavior $\xi_j$ is preferred over $\xi_i$.

We interpreted this ranking as a collection of pairwise comparisons, as in previous works \cite{brown2019extrapolating}. 
We adopted the Bradley-Terry preference model \cite{bradley1952rank} to model the probability that the user chooses behavior $\xi_j$ from the pair of behaviors $(\xi_i, \xi_j)$ \ndedit{based on the feature space mapping $\Phi$ (i.e., learned with CLEA or other self-supervised objectives)}: 

\vspace{-.5em}
\begin{equation}\label{eq:optimization_objective}
    P(\xi_i \prec \xi_j | R_H) = \frac{e^{R_H(\Phi(\xi_j))}}{e^{R_H(\Phi(\xi_i))} + e^{R_H(\Phi(\xi_j))}}
\end{equation}

To learn the user's reward function, $R_H$, we maximize the probability of all pairwise comparisons induced by the rankings the user performed. 
We construct a dataset containing all the users rankings, $\mathcal{D}_{pref.} = \{(Q_0, \sigma_0), (Q_1,\sigma_1), ...(Q_K,\sigma_K)\}$.
We minimize the total cross entropy loss summed over all pairwise comparisons across all rankings:

\vspace{-1em}
\begin{equation}\label{eq:reward_loss}
    \mathcal{L}(R_H) = \sum_{(Q,\sigma) \in \mathcal{D}_{pref}} \sum_{i=0}^{|Q|-1} \sum_{k=i+1}^{|Q|} -\log P(\xi_{\sigma(i)} \prec \xi_{\sigma(k)} | R_H)
\end{equation}

$R_H$ can be any computational model that can update its parameters to minimize a loss function. In this work, we used both neural networks and linear models to approximate $R_H$ to compare with prior works, however other techniques such as gaussian processes \cite{biyik2024active} are possible.

\section{Collecting Exploratory Actions from a Robot Customization Session}\label{sec:data_collection}

In this section, we describe our methodology for collecting user \ndedit{exploratory actions} in a free-form customization session involving a Kuri robot performing an item-finding task. 

 \begin{figure}
  \begin{center}
    \includegraphics[width=\linewidth]{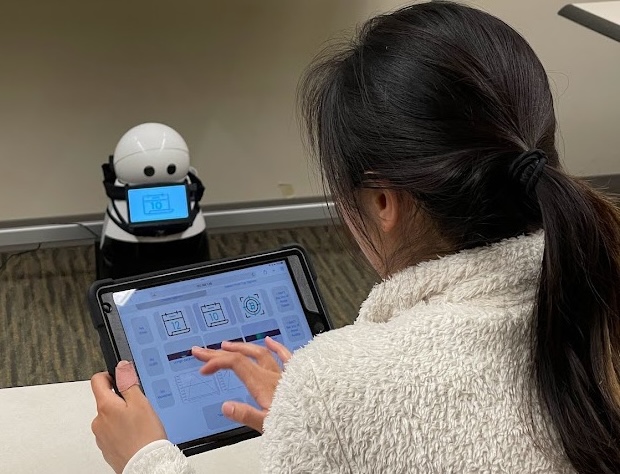}
  \end{center}
  \caption{\textbf{Customization session setup.} Participant designing signals for the modified Kuri robot using the query-based interface. }
  \label{fig:setup}
\end{figure}

 \begin{figure}
  \begin{center}
    \includegraphics[width=\linewidth]{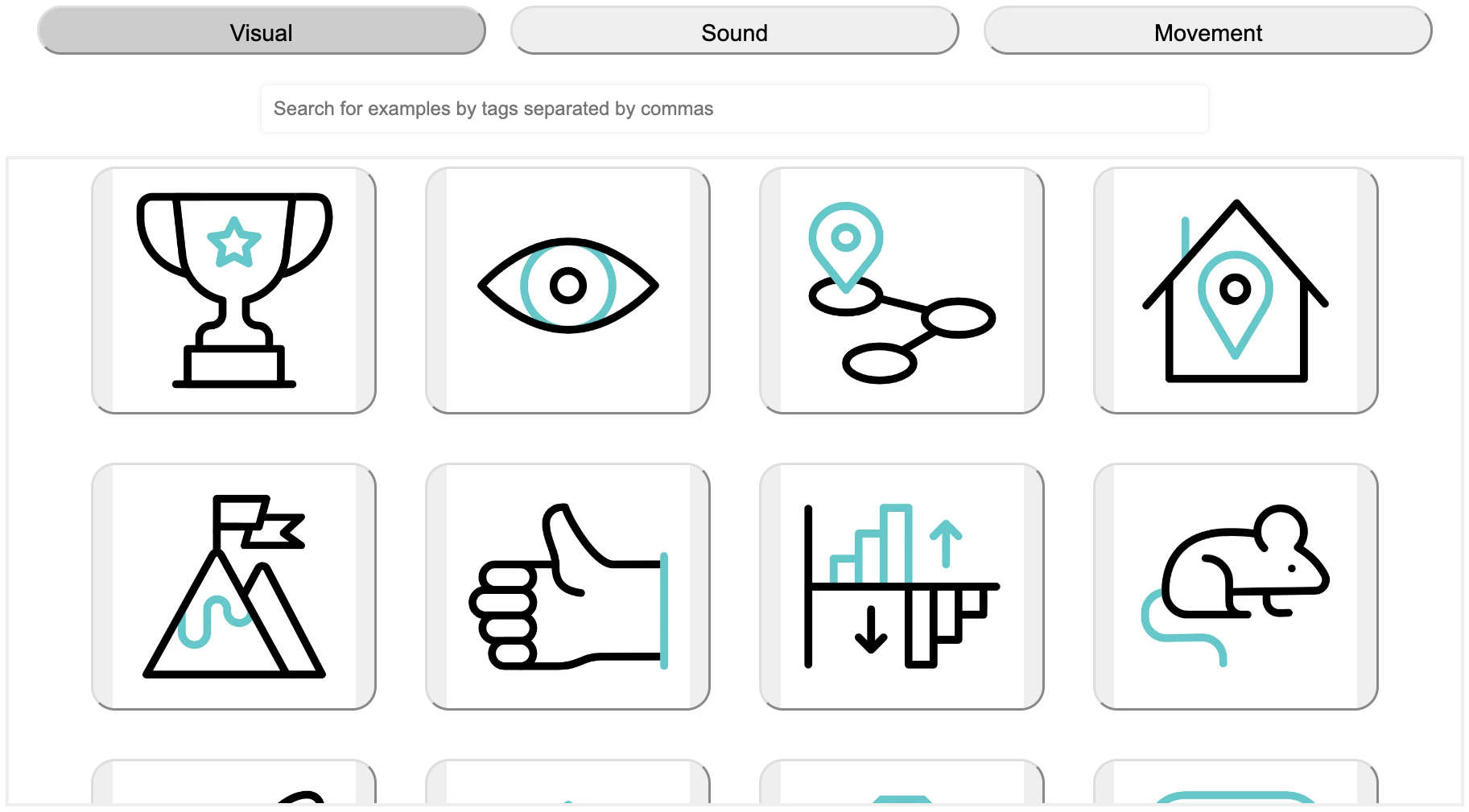}
  \end{center}
  \caption{\textbf{Exploratory search interface.} The exploratory search interface the participants used to design robot signals. Participants could explore visual, auditory, and kinetic robot behaviors \ndedit{by scrolling through the behavior options and by typing in search terms to filter results.}}
  \vspace{-2em}
  \label{fig:exploratory_interface}
\end{figure}

\subsection{Procedure}
We adapted a signal design task from previous work \cite{dennler2023rosid} to collect \textit{exploratory action} data from users designing multimodal signals for a robot to express the robot's state while assisting the user in an item-finding task.
Robot signaling presents a wide array of unconstrained and diverse user preferences, as opposed to purely functional tasks, where users tend to converge to a small set of effective robot behaviors \cite{sadigh2017active,biyik2024active}. Additionally, signaling does not require experimenters to ``engineer" user responses by telling them what their preference should be \cite{bobu2023sirl}.

We recruited participants to design signals for a robot that assists them in finding items around a room. 
This study received ethical approval from our university's Institutional Review Board.
Participants designed four state-expressive signals for a Mayfield Kuri robot \cite{kuri_robot}. We chose Kuri because it was designed to be a low-cost consumer product capable of natural interaction with non-expert users. For this study, we modified Kuri with an added screen and backpack to hold items, as shown in \autoref{fig:setup}. 
 
The participants were tasked with designing the following four signals: 
(1) idle, indicating that the robot is ready to receive a command; 
(2) searching, indicating that the robot is actively looking for a requested item; 
(3) has-item, indicating that the robot has the requested item in its backpack; and 
(4) has-information, indicating that the robot found the requested item but will need the user's assistance to procure the item.

Each signal consisted of three separate modalities: 
(1) \textbf{visual}: video played on the robot's screen; 
(2) \textbf{auditory}: sound played through the robot's speaker; and 
(3) \textbf{kinetic}: robot head movement.
These three modalities correspond to common data structures in robotics \cite{bobu2023sirl,cui2021empathic,yoo2024poe}: the visual modality is a sequence of images, the auditory modality is a spectrogram, and the kinetic modality is a sequence of joint angles. 
These robot behaviors pulled from a database of 5,192 videos, 867 sounds, and 2,125 head movements\footnote{\ndedit{The full dataset of behaviors is available to view on our project page: \href{https://interaction-lab.github.io/CLEA/}{https://interaction-lab.github.io/CLEA/}}}.

To engage in customizing the robot's signaling behaviors, users were presented with the RoSiD interface \cite{dennler2023design}. 
This interface allowed users to specify preferences for the robots in two ways. In the \textit{query-based} interaction for customizing robots, shown in \autoref{fig:setup}, participants were presented with a set of three options and chose their favorite among them.
In the \textit{exploratory search} interaction, shown in \autoref{fig:exploratory_interface}, participants were presented with up to one hundred options which they could filter, scroll through, and test on the physical robot.
\ndedit{We designed the \textit{exploratory search} interface to allow users to explore robot behaviors without having to repeatedly answer questions about rating the specific behaviors, which leads to user fatigue and decreased motivation \cite{schrum2023concerning}.}
Participants were allowed to freely choose either the \textit{query-based} interface or the \textit{exploratory search} interface to design signals for the robot. Participants were instructed to end the signal design when they were satisfied with their signal. 

After a participant designed all four signals, they interacted with the robot to complete an item-finding task, where the robot was piloted by an experimenter. 
The robot used the signals the participant designed to complete the interaction and allow the user to evaluate their design choices. 
Following the interaction, participants were interviewed to determine how they liked to design signals.
The interview data are outside of the scope for this paper. See Appendix \ref{appendix:demographics} for more details.

\subsection{Data Collection Results}
In total, 25 participants (10 women, 3 non-binary, 12 men; see Appendix \ref{appendix:demographics} for more demographic information) were recruited to design signals for the robot. Participants were compensated with US\$ 20 Amazon gift cards.

In total, participants spent an average of 415 seconds using the query-based interface and 654 seconds using the exploratory search interface. 
Each participant performed an average of 15.32 query interactions and 72.72 \textit{exploratory actions} across the four signals. \ndedit{Query interactions took an average of 27 seconds for a user to produce, compared to an average of 9 seconds for exploratory actions.}
We define \textit{exploratory actions} as the robot behaviors from the exploratory search interface that the user chose to evaluate on the real robot because they appeared appealing, compared to the behaviors that the user ignored because they seemed irrelevant. \ndedit{These exploratory actions encompass \textbf{all} actions the user took to explore robot behaviors, including examined behaviors that were seemingly unrelated to the specific signal. These unrelated explorations were generally actions users took to learn more about the robot's capabilities.}

We use the data we collected from all users performing exploratory actions to learn features of robot behaviors according to \autoref{eq:training_objective}. These features can be used for downstream preference learning tasks, such as behavior comparisons \cite{biyik2019asking, sadigh2017active, lu2022preference} or behavior rankings \cite{brown2019extrapolating, myers2022learning,chen2021learning}. \ndedit{To select hyperparameters, we used the data we collected from the \textit{query-based} interface. For more training details see Appendix \ref{appendix:training_details}}. \ndedit{Our previous work showed preliminary efficacy for CLEA by evaluating with data from the \textit{query-based} interface using leave-one-out cross-validation \cite{dennler2024using}. We expanded on those results by recruiting a new set of participants performing behavior rankings to evaluate CLEA on a new user population, detailed in the next section.}


\section{User Study Evaluation}\label{sec:experiments}

To evaluate the efficacy of learning feature spaces for robot behaviors, we conducted an experiment with a new set of participants. 
The participants ranked behaviors to generate individual datasets that we could use to quantitatively test different feature-learning algorithms.

\subsection{Manipulated Variables} 
To evaluate the effectiveness of learned features using automatically collected data, we evaluated seven total algorithms for learning feature spaces. The first baseline was: 
(1) \textbf{\textit{Random}}, a randomly-initialized neural network that projects each behavior to a vector. Random networks can be effective feature learners, as they cannot overfit to data or learn spurious correlations. 
(2) \textbf{\textit{Pretrained}}, a large pre-trained neural network to generate features. \ndedit{We used the vision-language foundation model X-CLIP \cite{ma2022x} to create features from videos of the visual and kinetic modalities. We used the audio foundation model AST \cite{gong2021ast} to create features for the auditory modality.}
We also evaluated two self-supervised baselines: 
(3) \textbf{\textit{AE}}, an autoencoder that uses a self-supervised loss to learn features that reconstruct the behavior, 
(4) \textbf{\textit{VAE}}, a variational autoencoder that uses a self-supervised loss to both reconstruct and standardize the distribution that the features come from. The AE and VAE methods use the latent space of these models as features. All of these self-supervised losses can also be combined with CLEA, so we evaluate the following as our proposed algorithms: 
(5) \textbf{\textit{CLEA}}, 
(6) \textbf{\textit{CLEA+AE}}, and 
(7) \textbf{\textit{CLEA+VAE}}. 
For all algorithms, we learned separate feature spaces for each of the three signal modalities: visual, auditory, and kinetic. The size of each feature space was a 128-dimension vector, which was sufficient to capture diverse preferences for complex behaviors \cite{chen2021learningelastic}. Additional information on the training processes is presented in Appendix \ref{appendix:training_details}.

\subsection{Procedure}\label{sec:eval_study}

\ndedit{To evaluate the generalizability of the features we learned, we collected ranking data from a separate set of 42 new participants (19 women, 4 non-binary, 19 men; more details in Appendix \ref{appendix:demographics}). }
Each participant completed ten behavior ranking trials for a particular modality and signal, with each ranking consisting of five robot behaviors.

The five behaviors we presented to the user for each ranking were selected based on the final customized signals in the customization session described in \autoref{sec:data_collection}, because previous work has shown that using other users' preferences is a good initialization for new users \cite{nikolaidis2015efficient,nemlekar2023transfer,dennler2023rosid,dennler2021personalizing}. 
To generate each of these five behaviors, we first randomly sampled a customized behavior from the customization session and then sampled one of the six feature-learning algorithms. 
Then, we calculated the behavior in the full database of behaviors that minimized the feature distance to the custom behavior we sampled, according to the feature space we sampled.
 
In order to fully evaluate the proposed algorithms, we must know the user's overall favorite behavior to use as a ground-truth preference. To achieve this, the fifth and tenth ranking used the top-ranked signals from the previous ranking trials to create a ``super ranking". The highest-ranked behavior in the final ranking trial represents the participant's overall favorite behavior.

\subsection{Hypotheses}  
A survey by Bobu et al. \cite{bobu2024aligning} identified four criteria that constitute good representations for downstream preference-learning tasks: 
(1) \textbf{\textit{Completeness}}, the ability of a representation to capture a user's true preferences,
(2) \textbf{\textit{Simplicity}}, the ability to recover user preferences from linear transformations of the representations,
(3) \textbf{\textit{Minimality}}, the ability of a representation to exist in low-dimensional spaces, and
(4) \ndedit{\textbf{\textit{Explainability}}, the ability of a representation to be compatible with existing explainability tools.} We adopt this framework for our analysis and present our additional results in Appendix \ref{appendix:additional_results}.

We evaluate these four criteria with the data collected from the 42 participants in \autoref{sec:eval_study}.
We split the data from each participant into 70\% for training our reward models, and 30\% for evaluating our reward models.

Based on this framework, we tested four hypotheses comparing CLEA features to self-supervised features:
\begin{enumerate}[label={\bf (H\arabic*)}, left=0em]
    \item Exploratory actions reflect user preferences, so the most \textbf{complete} features will leverage CLEA;
    
    \item Exploratory actions align with preference teaching tasks, so the most \textbf{simple} features will leverage CLEA; 
    \item Exploratory actions efficiently express user preferences, so the most \textbf{minimal} features will leverage CLEA; and
    \item Exploratory actions are semantically meaningful, so the most \textbf{explainable} features will leverage CLEA.
\end{enumerate}

\subsection{Results}\label{sec:results}

\begin{figure}
  \begin{center}
    \includegraphics[width=\linewidth]{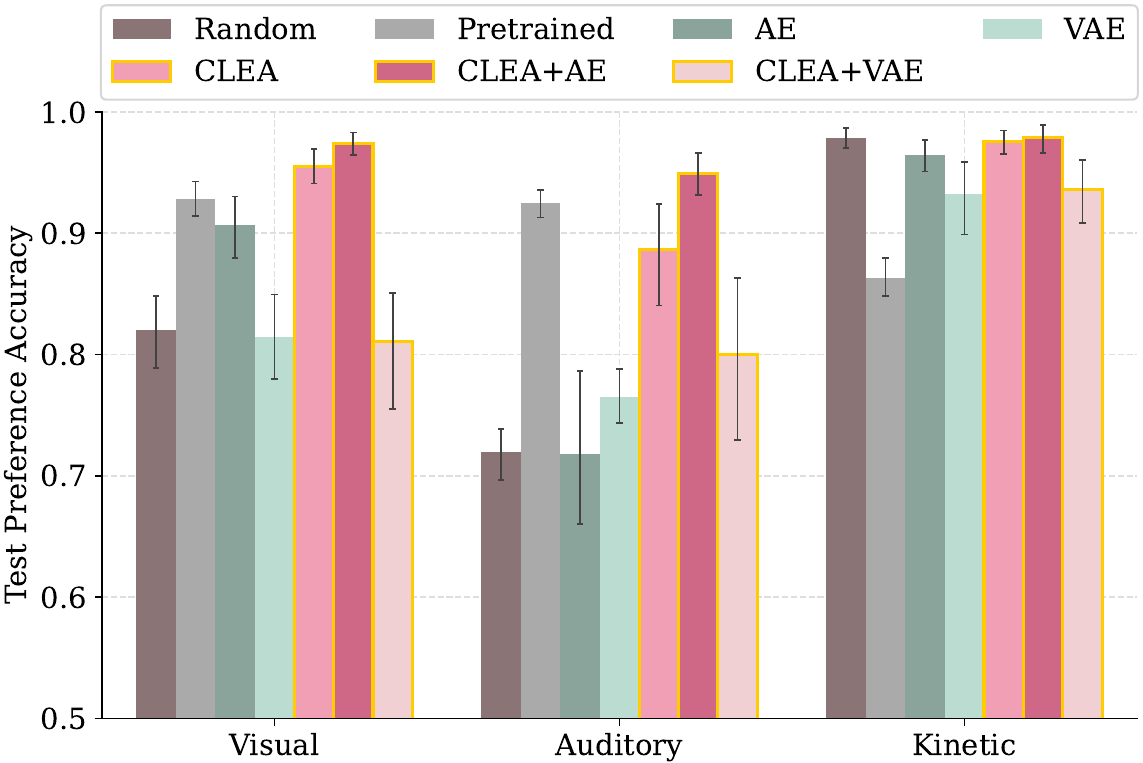}
  \end{center}
  \vspace{-1em}
  \caption{\textbf{Completeness results.} Across three modalities, feature spaces using CLEA are able to accurately predict user preferences. Error bars show mean standard error across participants.}
  \vspace{-1em}
  \label{fig:Completeness}
\end{figure}

\begin{table*}[t]
\caption{\textbf{Simplicity results.} For each modality, we found the area under the curve (AUC) of the alignment metric over 100 pairwise comparisons across feature dimensionalities. Asterisks indicate best-performing algorithm within each dimension (all $p <.05$).} \label{tbl:minimality}

\resizebox{\textwidth}{!}{%

\begin{tabular}{rccccc|ccccc|ccccc}

\midrule

\multicolumn{1}{l}{} & \multicolumn{5}{c|}{Visual}                                                              & \multicolumn{5}{c|}{Auditory}                                                           & \multicolumn{5}{c}{Kinetic}                                                           \\
Dimension                 & 8               & 16              & 32              & 64              & 128             & 8               & 16              & 32             & 64              & 128             & 8              & 16              & 32             & 64              & 128             \\ \midrule
Random               & .005                  & .024                   & .018                   & .013                   & .004                    & -.001                 & .134                   & -.001                  & .000                   & .000                    & .003                  & .179                   & .091                   & .187                   & .272                    \\

\ndedit{Pretrained} & \ndedit{.127} & \ndedit{.056} & \ndedit{.047} & \ndedit{.042} & \ndedit{.035}  & \ndedit{.267} & \ndedit{.022} & \ndedit{.021} & \ndedit{.025} & \ndedit{.025} & \ndedit{.051} & \ndedit{.042} & \ndedit{.051} & \ndedit{.057} & \ndedit{.055}\\

AE                   & .024                  & .014                   & .011                   & .014                   & .007                    & .038                  & .065                   & .042                   & .080                   & .015                    & .014                  & .227                   & \textbf{.330*}         & .321                   & .154                    \\
VAE                  & .269                  & \textbf{.335*}         & .247                   & .180                   & .033                    & .234                  & .174                   & .117                   & .083                   & .077                    & .207                  & .251                   & .192                   & .203                   & .346                    \\

\midrule
CLEA                 & .012                  & .261                   & .245                   & .090                   & .044                    & .058                  & .002                   & .142                   & .113                   & .046                    & \textbf{.284*}        & .224                   & .255                   & .345                   & .217                    \\
CLEA+AE              & \textbf{.315*}        & .219                   & .330                   & .163                   & \textbf{.275*}          & .260                  & .023                   & .141                   & .015                   & .140                    & .079                  & .208                   & .192                   & .207                   & .147                    \\
CLEA+VAE             & .196                  & .295                   & \textbf{.376*}         & \textbf{.293*}         & .147                    & \textbf{.438*}        & \textbf{.343*}         & \textbf{.236*}         & \textbf{.198*}         & \textbf{.175*}          & .009                  & \textbf{.260*}         & .165                   & \textbf{.373*}         & \textbf{.377*} \\ \midrule
\end{tabular}%
}

\vspace{-1em}
\end{table*}

\begin{figure*}
    \centering
    \includegraphics[width=.96\textwidth]{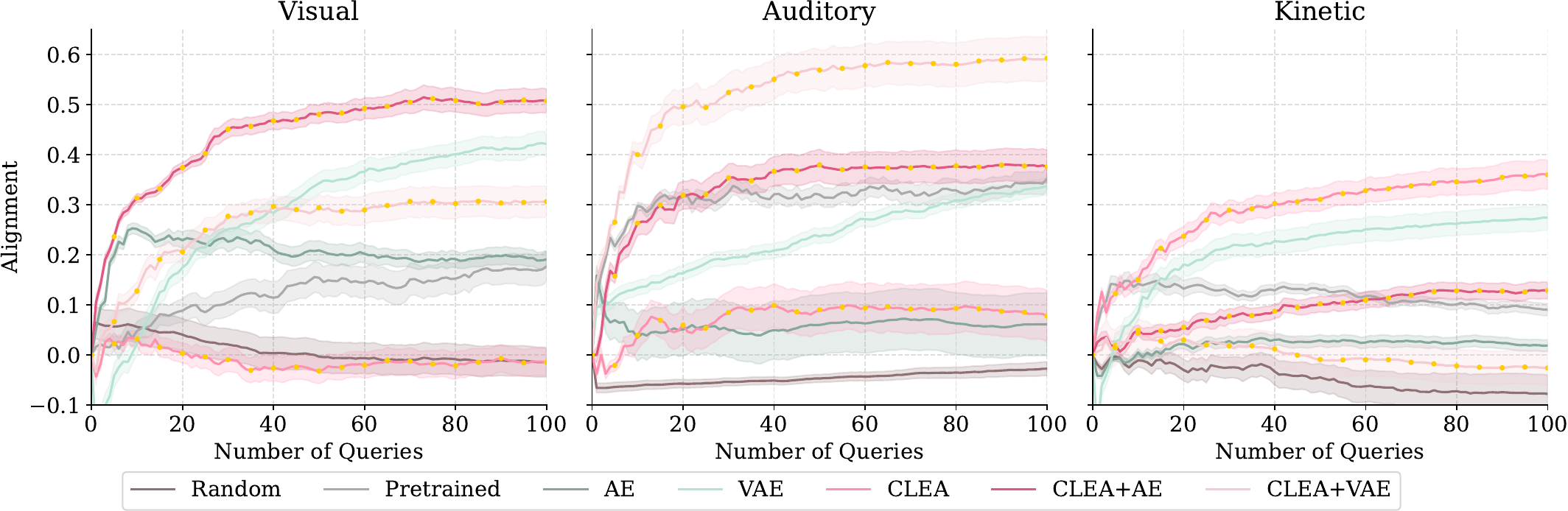}
    \caption{\textbf{Minimality results.} Alignment of a linear reward model across numbers of pairwise comparisons for the smallest sized feature space. Shaded region indicates mean standard error.}
    \label{fig:simplicity}
    \vspace{-1.5em}
\end{figure*}

\textbf{Evaluating Completeness.} Completeness refers to the feature's ability to capture all relevant information to understand how a user ranks robot behaviors.
\ndedit{To evaluate completeness, we aimed to learn a \textbf{neural network} reward model that can accurately model the choices participants made during the ranking experiment. We quantitatively measured completeness with the \textit{test preference accuracy} (TPA) metric \cite{bobu2023sirl}, which measures the accuracy of a trained model to predict the user's choice in an unseen test set.
To predict user choices, we used the 128-dimensional feature spaces for the six algorithms as input to a neural network that estimated the participant's internal reward. 
This reward network that consisted of two fully connected layers with hidden dimensions of 256 units to output a single value.}
The training objective maximized the probabilities of the selected behaviors in the training set using \autoref{eq:optimization_objective}.

The TPA of all three modalities is shown in \autoref{fig:Completeness}. We show that CLEA+AE had the highest TPA in modeling participant's choices in the Visual and Auditory modality, determined by t-tests (all $p<.05$). 
In the Kinetic modality, CLEA+AE, CLEA, and Random were tied for the highest TPA, but had a higher TPA than the other methods ($p<.05$).  CLEA features contain \textbf{\textit{complete}} information to model user preferences, supporting \textbf{H1}. For an extended statistical anaylsis, see Appendix \ref{appendix:completeness_stats}.

\textbf{Evaluating Simplicity and Minimality.} \ndedit{A feature space is \textbf{\textit{simple}} if it can model a user's preference with a \textbf{linear} reward model, and it is \textbf{\textit{minimal}} if the dimensionality of the feature space that is used as input to this reward model is small while still accurately capturing user preferences \cite{bobu2024aligning}.}
Simple linear reward models are practically useful compared to complex neural network reward models because linear models are easier to store, interpret, and compare \cite{reed2022self}.
\ndedit{To quantify both minimality and simplicity, we used the area under the curve of alignment (AUC Alignment) \cite{biyik2019asking,sadigh2017active,myers2022learning} over 100 pairwise queries.
We evaluated AUC Alignment across feature spaces of five dimensions: 8, 16, 32, 64, and 128.
The simple linear model to estimate a user's reward was described as $R_H(\xi) = \omega \cdot \Phi(\xi)$.
We estimated $\omega$ using Bayesian inverse reward learning, as in previous works \cite{biyik2019asking,sadigh2017active,brown2019extrapolating}.}

To calculate AUC Alignment, 
we sequentially updated the estimate of the user's preference $\omega_{est.}$ after observing each ranking action they made, decomposed into pairwise queries. We calculated the alignment of the user's true preference $\omega_{true}$ and estimated preference, $\omega_{est.}$, following the equation, $\frac{1}{M}\sum_{\omega_{est.}\sim\Omega} \frac{\omega_{true}\cdot\omega_{est.}}{||\omega_{true}||_2\cdot||\omega_{est.}||_2} $, from prior work \cite{biyik2019asking,sadigh2017active}. 
We set the user's true preference, $\omega_{true}$, as the vector corresponding to the user's top-ranked signal. 
\ndedit{We used AUC Alignment \cite{biyik2019asking,sadigh2017active,myers2022learning} over the number of queries as the metric to assess simplicity and minimality.} 
A higher AUC Alignment indicates that we learned the user's preference more accurately and with fewer queries. We show the alignment curve in \autoref{fig:simplicity}.

To evaluate \textbf{\textit{simplicity}}, we compared AUC Alignment of our simple model across all five feature space dimensions to show that a simple model effectively models preferences for all dimensions; the results are shown in \autoref{tbl:minimality}. 
We observed that across all modalities, a CLEA-based feature space has the highest AUC Alignment in 13 of the 15 experiments, with CLEA+VAE being the best on 10 of these experiments.
Overall, training with the CLEA objective resulted in \textbf{\textit{simple}} representations that were useful across different sized dimensions, supporting \textbf{H2}. 
Further analysis is in Appendix \ref{appendix:simple_stats}.

To evaluate \textbf{\textit{minimality}}, we compared AUC Alignment for only the 8-dimensional feature space to determine if CLEA can model user preferences for low-dimensional feature spaces.
The results are shown in \autoref{fig:simplicity}.
We found that CLEA+AE has the highest AUC Alignment in the Visual modality, CLAE+VAE has the highest AUC Alignment in the Auditory modality, and CLEA has the highest AUC Alignment in the Kinetic modality.
We conclude that using a loss function that includes the CLEA objective results in features that \textbf{\textit{minimally}} elicit user preferences, supporting \textbf{H3}. For additional analysis see Appendix 
\ref{appendix:minimal_stats}.

\begin{figure}
  \begin{center}
    \includegraphics[width=.92\linewidth]{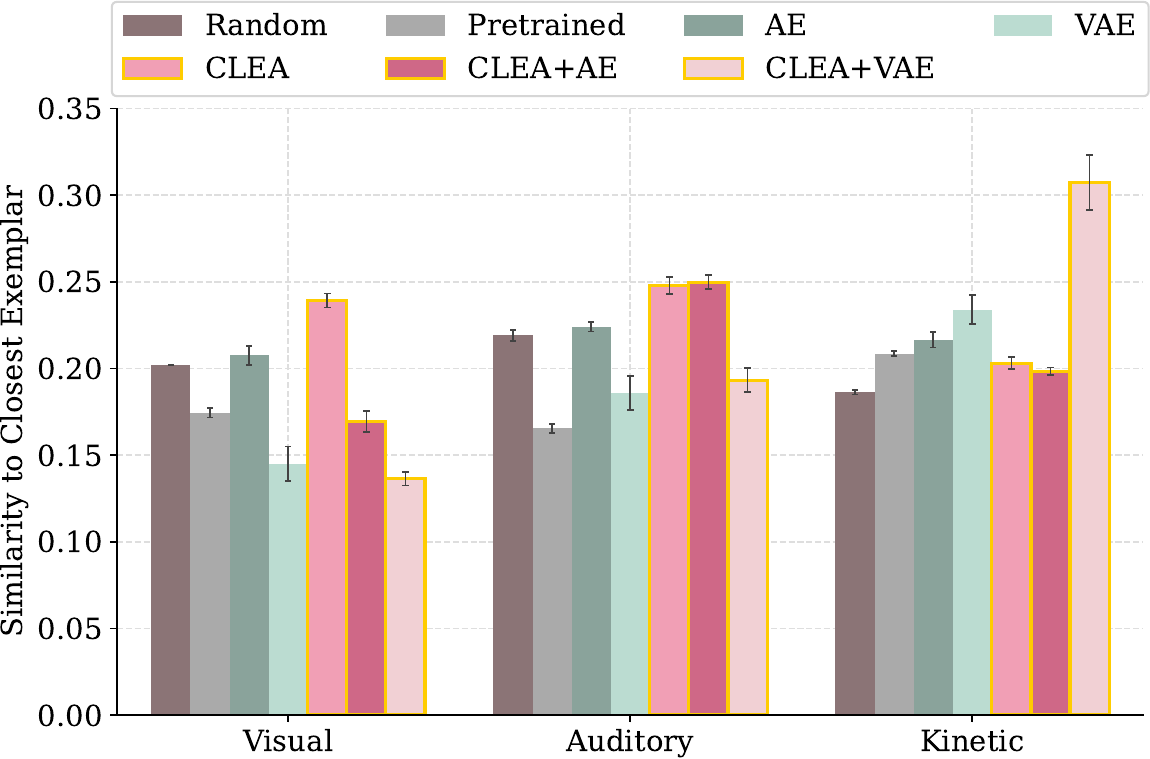}
  \end{center}
  \vspace{-1em}
  \caption{\textbf{Explainability Results.} \ndedit{We examined the similarity between top-ranked signals from the ranking user study and the nearest exemplar signal from the robot customization session. We found that CLEA-based feature spaces have higher similarities, facilitating explanation by example. }
  }
  \vspace{-1em}
  \label{fig:Explainability}
\end{figure}

\textbf{Evaluating Explainability.}
\ndedit{A common explainability technique for neural systems is explanation by example \cite{caruana1999case,papernot2018deep}. In this framing, the user is presented with training examples, called exemplars, that are near unseen test samples to provide an interpretation of learned feature spaces. This approach has been used to explain feature spaces used for clustering \cite{been2014the}, image recognition \cite{hendricks2016generating}, and language \cite{mikolov2013distributed}. In this work, we quantitatively evaluated explainability of the feature space by measuring the cosine similarity of the top-ranked signals from the ranking study (\autoref{sec:eval_study}) to their nearest exemplar from the customization session (\autoref{sec:data_collection}). We show the similarity scores for each modality in \autoref{fig:Explainability}.}


\ndedit{We found that the CLEA feature space resulted in the highest similarity between top-ranked signals and the nearest exemplars for both the visual and auditory modality. The CLEA+VAE feature space demonstrated the highest similarity in the Kinetic modality. These results indicate that CLEA-based feature spaces are more conducive to explanation by example than self-supervised feature spaces, thus}
CLEA results in the most \textbf{\textit{explainable}} feature spaces, supporting \textbf{H4}.



\section{Discussion and Limitations}
Our results demonstrate the efficacy of learning useful features for eliciting preferences by leveraging data from natural user interface interaction. 
By incorporating exploratory search concepts into interfaces for teaching robots, we can scale data collection while also providing users with an intrinsically motivating task. 
We performed evaluations across three modalities of state-expressive signals in robots---visual, auditory, and kinetic---and found that using CLEA significantly increased performance in all modalities and evaluation criteria.

CLEA can be readily combined with any algorithm that learns feature spaces for eliciting preferred robot behaviors by learning a lower-dimensional embedding of the behaviors. 
We demonstrated CLEA's use with self-supervised losses; it can also be used with other methods of learning feature spaces such as trajectory similarity queries \cite{bobu2023sirl}, multi-task learning \cite{nishi2020fine,yamada2022task}, and labelled behaviors \cite{nair2022learning,sontakke2024roboclip}. 

We developed an algorithm for learning feature spaces using exploratory search actions. 
This requires users to be able to briefly review many behaviors at a high level before deciding which of these behaviors might be relevant and warrant further exploration. 
We used visual summaries for all modalities--a video frame for the visual modality, a spectrogram for the auditory modality, and a graph of joint angles for the kinetic modality. 
These were interpretable by the participants, but they may not be the most effective means of representing the underlying behaviors. 
Users may be better at interpreting natural language descriptions \cite{li2017person}, tags that describe the behavior \cite{chang2019searchlens}, or animated gifs \cite{bakhshi2016fast}. 
Future work can explore how robot behaviors can be summarized to non-expert users in ways that allow them to most efficiently search through robot behavior options. 
Understandable summaries are especially needed for users to perform exploratory search with more complex robot behaviors, such as different gaits for quadruped robots or dexterous manipulation skills for high degree of freedom manipulators. 

\ndedit{The robot behaviors that users explored often appeared unrelated to the user's personal preferences, but these seemingly random explorations were still indicative of other users' preferences, as demonstrated by the transferability of CLEA to distinct populations.}
We also assumed that users would be motivated to perform this exploratory search since they were unfamiliar with what the robot was capable of \cite{marchionini2006exploratory}. 
If users are already familiar with a particular robot, they may not be motivated to perform exploratory actions because they have already found their preferred robot behaviors. This familiarity could decrease the efficacy of CLEA as a framework for learning feature spaces in already adopted robots. 

We also found that additional loss terms, such as the reconstruction term and the KL-divergence term, were only sometimes helpful for training with CLEA, depending on the underlying data structures used. 
We found that the additional reconstruction loss and KL-Divergence loss were often helpful in video and sound data structures but could hinder preference learning for joint state sequences. 
This effect is due to the social interpretations interfering with these additional terms. 
For joint state behaviors, gestures portraying fear and excitement have very similar joint states, but vastly different social interpretations. 
CLEA aims to separate these features, while the reconstruction term brings them together. 
While the optimal loss terms can only be determined experimentally, we presented a set of metrics for evaluation that can systematically determine these terms, in accordance with the qualities of good feature spaces \cite{bobu2024aligning}.

\textbf{Conclusion.} We present \textit{contrastive learning from exploratory actions} (CLEA), an algorithm to leverage a novel data source of interactions that users automatically perform when teaching robot systems. 
We showed that CLEA can be used to learn feature spaces that reflect underlying personal preferences, represent robot behaviors in low-dimensional vectors, quickly elicit user preferences, and are explainable.

\section*{Acknowledgments}

\fontsize{8.5}{9.7}\selectfont
This work was \textbf{funded} in part by an \textbf{Amazon Research Award} and a \textbf{National Science Foundation Graduate Research Fellowship Award} (\#DGE-1842487). The visual components users selected from to create the robot's signals were sourced from \href{https://www.flaticon.com/animated-icons}{Flaticon.com}.

\fontsize{10}{12}\selectfont

\newpage
\balance
\bibliographystyle{IEEEtran}
\bibliography{main}

\clearpage

\appendix

\section*{Additional Demographic and Study Information} \label{appendix:demographics}

\ndedit{In Appendices A and B, we report additional study details and demographic information. Appendix A discusses the customization session from \autoref{sec:data_collection}, and Appendix B discusses the evaluation study from \autoref{sec:eval_study}.}

\subsection{Robot Customization Session}
\textbf{Additional Demographic Information.} 
A total of 25 participants were part of the study, with ages ranging from 19 to 43 (median 25); participants self-declared as men (13), women (10), and genderqueer, non-binary, or declined to state (3, aggregated for privacy; some participants belonged to multiple groups). 
We recruited 13 participants who self-identified as LGBTQ+. Participants were Asian (13), Black (2), Latino (5), and White (6); some participants belonged to multiple groups. 
All participants were able to create signals they liked for all four categories, and all successfully interacted with the robot to collect all the items in the word search task.

\textbf{Additional Procedure Information.}
Participants were recruited from the local university student population through email, flyers, and word-of-mouth. 
\ndedit{The study took place in a conference room with a kitchen to reflect a realistic living environment. Participants entering the study were brought to a table in the middle of the room, with a clear view of a Kuri robot that was modified to have a screen and backpack (see \autoref{fig:setup} for a visual of the robot). }

\ndedit{First, the experimenter provided a ten-minute explanation of the study. In this explanation, participants were first introduced to the item-finding task, and were described each of the four signals in detail (\textit{idle}, \textit{searching},\textit{ has-item}, and \textit{has-info}). The experimenter introduced the participant to the RoSiD interface and described how to use each part of the interface. Once the introduction concluded, participants were instructed to design each of the four signals in a randomized and counterbalanced order.}

\ndedit{For each of the signals, participants were allowed to design any signal that they liked. There was no time limit, and participants were able to continue customizing until they reached a finalized signal. Once they finished designing, they were instructed to tell the experimenter they were ready to move to the next signal. When participants completed designing all four signals, the participant used the robot in its intended use case of finding items.}

\ndedit{The item-finding task aimed to simulate using the robot to find items while being distracted by other tasks. We achieved this by having the users engage in a word search. There were ten total words to find, but only seven of these words were listed on the word search sheet. To find the other three words and complete the word search sheet, users had to interact with the robot to find items around the room. When the robot returned these items to the user, the item had the word to find in the word search printed on a label physically attached to the item. The three items were: a stapler (with the label ``interactive"), a salt shaker (with the label ``kinematics"), and a doorstop (with the label ``haptic"). The salt-shaker and doorstop were items that Kuri used the \textit{has-item} signal for, because they were small enough to fit in the backpack on Kuri. The stapler was too big to fit in Kuri's backpack, and thus Kuri used the \textit{has-info} signal to have the user stand up and walk over to the Kuri robot to pick up the item. The stapler was placed on a counter behind the participant, out of view from the table that the participant was facing.}

\ndedit{Following the interaction, participants filled out the system usability scale. The experimenter then performed a semi-structured interview with the participant to understand their opinions on the design process. Participants then completed the study, and were compensated with an Amazon Gift Card sent to their email.}

\subsection{Preference Evaluation Study}
Participants were recruited from the local university student population through email, flyers, and word-of-mouth. A total of 42 participants were part of the study, with ages that ranged from 18 to 32 (median 24); participants self-declared as men (19), women (19), and genderqueer, non-binary, or declined to state (4, aggregated for privacy; some participants belonged to multiple groups). 
There were 17 participants that self-identified as LGBTQ+.
Participants were Asian (24), Black (1), Latino (7), Middle Eastern (3), and White (11) (some participants belonged to multiple groups). 
Participants rated their median familiarity with robotics as a 3 out of 9; a score of 1 corresponded with the term ``novice" and a score of 9 corresponded with the term ``expert".

\section*{Additional Training Details}
To encourage reproducibility, we provide the specifics of our training experiments in Appendices C and D. Appendix C discusses the feature learning models, and Appendix D discusses the reward learning models.

\subsection{Training Feature Learning Models}\label{appendix:training_details}
We used the following encoder architectures for each modality. We used the transposed architecture for all self-supervised methods that required a decoder.

\begin{algorithm}[t]
\caption{Contrastive Learning From Exploratory Actions}\label{alg:CLEA}
\textbf{Given} a list of robot trajectory datasets that all users saw over the course of the signal design process separated into explored and ignored data, $ D = \{(\mathcal{D}_i^{ex.},\mathcal{D}_i^{ig.})_{i=0}^N \}$, a learnable model that generates trajectory features, $\Phi$, and a hyperparameter for the contrastive margin, $\alpha$\;

\textbf{Initialize} $\Phi$ to a random state (or to a pretrained network)\; 

\While{not converged}{
$(\mathcal{D}^{ex.}, \mathcal{D}^{ig.}) \gets$ sample item from D\;

\tcp{Sample anchor and positive from explored data}
\If{$\text{Uniform}(0,1) < 0.5$}{
    $\xi_A \sim \mathcal{D}^{ex.}, \xi_P \sim \mathcal{D}^{ex.}, \xi_N \sim \mathcal{D}^{ig.}$\;
    }
    \tcp{Sample anchor and positive from ignored data}
\Else{
$\xi_A \sim \mathcal{D}^{ig.}, \xi_P \sim \mathcal{D}^{ig.}, \xi_N \sim \mathcal{D}^{ex.}$\;
}

$\mathcal{L}_1 = max(||\Phi(\xi_A) - \Phi(\xi_P)||^2_2 - || \Phi(\xi_A) - \Phi(\xi_N)||^2_2+ \alpha, 0)$\;
$\mathcal{L}_2 = max(||\Phi(\xi_P) - \Phi(\xi_A)||^2_2 - || \Phi(\xi_P) - \Phi(\xi_N)||^2_2+ \alpha, 0)$\;
update parameters of $\Phi$ to minimize $\mathcal{L}_1 + \mathcal{L}_2$

}

\end{algorithm}

\begin{itemize}
    \item Visual: the visual modality used a convolutional architecture that consisted of kernels with sizes: $(16,16), (8,8), (4,4)$ followed by a three-layer MLP with hidden size 256. Each convolutional layer was followed by a batch norm and a leaky ReLU activation. Each MLP layer except the last was followed by a ReLU activation   
    \item Auditory: the auditory modality used a convolutional architecture that consisted of kernels with sizes $(16,16), (8,8), (4,4)$ followed by a three-layer MLP with hidden size 256. Each convolutional layer was followed by a batch norm and a leaky ReLU activation. Each MLP layer except the last was followed by a ReLU activation   
    \item Kinetic: the kinetic modality used a recurrent architecture consisting of a bidirectional 2-layer GRU with a size 64 dimension hidden state.
\end{itemize}

We provide the pseudocode to train a network with the CLEA objective in \autoref{alg:CLEA}. For all feature learning models we used the Adam optimizer with default learning rates. All feature learning models also used a batch size of 128. We trained a separate feature learning model for each signal that users designed for.

We selected hyperparameters for the networks using the query data collected in the robot customization session (\autoref{sec:data_collection}) as a validation set. All our methods had three possible terms in our loss function: the contrastive loss that we formulated, a reconstruction loss, or a KL-divergence loss comparing the batch distribution to a unit multivariate normal distribution. Only the contrastive loss and the KL-Divergence loss had tunable parameters. To select the margin for the contrastive loss, we performed a parameter sweep over $\alpha \in [.01,.1,.5,.9,2,5,10.]$. The highest performing $\alpha$ values were $\alpha = .1$ for the visual modality, $\alpha = .1$ for the auditory modality, and $\alpha = 2$ for the kinetic modality. To select the regularization term for the VAE, we performed a parameter sweep over $\beta \in [.01,.1,.5,.9,2,5,10.]$. The highest-performing $\beta$ values were $\beta = 1$ for the visual modality, $\beta = 10$ for the auditory modality, and $\beta = 10$ for the kinetic modality.

\begin{figure*}[!ht]
    \centering
    \includegraphics[width=\textwidth]{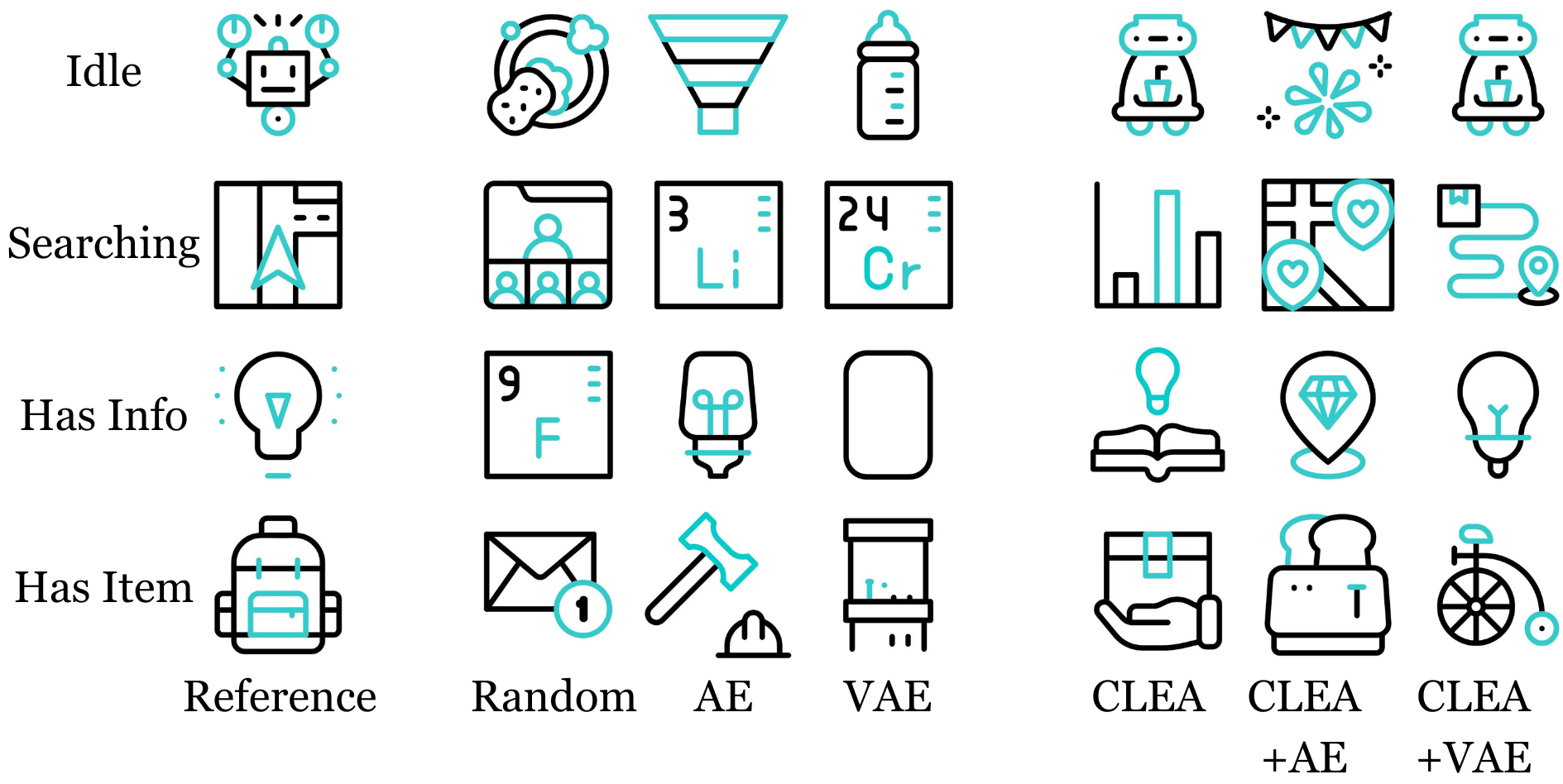}
    \caption{Qualitative results. The leftmost image shows a reference image that users actually selected when designing signals for the robot. The other images show the next most similar image in the embedding space for each method. CLEA-based methods show more semantic similarity to the reference image than self-supervised approaches.}
    \label{fig:qualitative}
\end{figure*}

\subsection{Reward Training}\label{sec:reward_training}

We learned two forms of reward functions to evaluate CLEA. The first was a reward network that we evaluated with predicted choice accuracy, similar to Bobu et al. \cite{bobu2023sirl}. The second form of reward was a linear transformation of the features as in Bıyık et al. \cite{biyik2019asking}.

\textbf{Neural Network Reward.} We used the same reward network for all modalities. The network takes as input a $d$-dimensional vector. The network itself consists of two fully connected layers, each with 256 hidden units. Each layer is followed by a ReLU nonlinearity. We trained the network using cross entropy loss (\autoref{eq:reward_loss}).

We additionally encouraged the learned reward to not be too large by placing a L2-norm regularization on the predicted rewards, with a weight of $.01$ for all reward networks. We trained all reward networks for 60 epochs using the default settings for the Adam optimizer and a batch size of 16.

\textbf{Linear Reward.} The linear reward models as user's reward function as $R_H(\xi) = \omega \cdot \Phi(\xi)$, where the user's specific preference is represented by $\omega$. We learned a user's $\omega$ through pairwise choices. We adopted the Bradley-Terry preference model to model the probability of a user choosing $\xi_k$ from a query $Q=\{\xi_0,\xi_1,...\xi_N\}$: 

\begin{equation}
    P(\xi_k | Q, \omega) = \frac{e^{\omega \cdot \Phi(\xi_k)}}{\sum_{i=0}^N e^{\omega \cdot \Phi(\xi_i)}}
\end{equation}

To learn $\omega$, we apply Bayes' rule.

\begin{equation}
    P(\omega | Q, \xi_k) \propto P(\xi_k | Q, \omega) \cdot P(\omega)
\end{equation}

We assume a prior $\omega$ of a uniformly distributed unit ball, as in prior works \cite{sadigh2017active}.  We update our posterior after every observed choice to estimate the user's $\omega$ using Monte Carlo methods.

\section*{Extended Statistical Analysis}
\ndedit{In Appendices E--H, we report the full statistical analysis we performed across algorithms to report effect sizes so that others may use these for power analyses. Appendix E discusses completeness, Appendix F discusses simplicity, Appendix G discusses minimality, and Appendix H discusses explainability.} 

\subsection{Completeness}\label{appendix:completeness_stats}
We statistically evaluated the empirical completeness results from \autoref{sec:results}. In that section, we evaluated \textit{test-preference accuracy} (TPA) for each of the seven algorithms (Random, Pretrained, AE, VAE, CLEA, CLEA+AE, and CLEA+VAE).
Our analysis used a repeated measures ANOVA for each modality, using TPA as the dependent measure and algorithm as the within-subjects factor. 
All assumptions were met. 
The choice of algorithm is significant for all modalities (visual: $p<.001, \eta^2=.582$; auditory: $p<.001, \eta^2=.532$; kinetic: $p=.008, \eta^2=.247$). 
We performed post-hoc analysis using paired t-tests with a Bonferroni correction. 
Our analysis revealed that CLEA and CLEA+AE outperformed all other algorithms in the visual modality (all $p_{corr.}<.05$) and CLEA+AE outperformed all algorithms in the auditory modality (all $p_{corr.} < .05$). 
There were no significant differences between algorithms in the kinetic modality; Random, CLEA, and CLEA+AE empirically performed the highest.

\subsection{Simplicity}\label{appendix:simple_stats}

We statistically evaluated the empirical simplicity results from \autoref{sec:results}. In that section, we evaluated \textit{Area Under the Alignment Curve} (AUC Alignment) for each of the seven algorithms across each of the five feature space dimensions (8, 16, 32, 64, and 128).
We performed a two-way repeated measures ANOVA with dimension and algorithm as within-subjects factors and AUC Alignment as the dependent measure. 
The feature space dimension and the feature learning algorithm were the within-subjects factor. 
We found that the choice of algorithm was significant for all modalities (visual: $p<.001,\eta^2=.909$; auditory: $p<.001,\eta^2=.502$; kinetic: $p<.001, \eta_p^2=.530$). 
We used pairwise t-tests with Bonferroni corrections to assess all pairwise comparisons between the algorithms for learning features.

For the visual modality, we found that CLEA+AE features were the best on average, significantly outperforming all other features across feature space dimensions (all $p_{corr.}<.001$). 
For the auditory modality, CLEA+VAE features were the best on average, significantly outperformed all algorithms across feature space dimensions (all $p_{corr.}<.001$). 
For the kinetic modality, CLEA features outperformed all algorithms across feature space dimensions except AE features (all other $p<.001$), however we note that CLEA has much higher performance than AE for lower-dimensional feature spaces. 

\subsection{Minimality}\label{appendix:minimal_stats}
For minimality, we evaluated AUC Alignment across the seven algorithms for only the 8-dimensional feature space. We performed a repeated measures ANOVA using AUC as a dependent measure and the algorithm as the within-subjects factor and determined that the choice of algorithm was significant (visual: $p<.001,\eta^2=.776$; auditory: $p<.001,\eta_p^2=.685$; kinetic: $p<.001, \eta_p^2=.658$) we used paired t-tests with 
Bonferroni correction to assess all pairwise comparisons. 
In the visual modality, CLEA+AE significantly outperformed all other algorithms (all $p_{corr.}<.001$). 
In the auditory modality, CLEA+VAE significantly outperformed all other algorithms (all $p_{corr.}<.001$). 
In the kinetic modality CLEA outperformed all other algorithms (all $p_{corr.}<.003$).

\subsection{Explainability}

We evaluated explainability with the cosine similarity between users' top ranked signals from \autoref{sec:eval_study} and the nearest user-designed exemplar signal from \autoref{sec:data_collection}. We evaluated the cosine similarity for the feature spaces generated by each of the seven algorithms. We evaluated significance using a one-way ANOVA with cosine similarity as the dependent measure and algorithm as the between subjects factor---each participant had only one top-ranked signal, so there was no repeated measure. All ANOVA assumptions were met. The choice of algorithm was significant for all modalities (visual: $p<.001$, $\eta^2=.362$; auditory: $p<.001$, $\eta^2=.295$; kinetic: $p<.001$, $\eta^2=.281$). We used pair-wise t-tests with Bonferroni corrections to assess pairwise differences.

For the visual modality, CLEA feature spaces produced the highest cosine similarity values over all other methods (all $P<.001$). In the auditory modality, CLEA and CLEA+AE produced the highest cosine similarity values over all other methods (all $p<.001$). In the kinetic modality, CLEA+VAE produced the highest cosine similarity values over all other methods (all $p<.01$)


\section*{Additional Experiments} \label{appendix:additional_results}

Appendices I--L show the results of the four additional experiments to evaluate CLEA. Appendix I qualitatively examines the structure of the feature space; Appendix J examines each method's robustness to injected noise; Appendix K examines the effect of time-based reweighting of exploratory actions to capture how users learn about the robot over time; and Appendix L examines directly learning the user's reward from from the raw robot behaviors without learning a feature space.

\subsection{Qualitative Results}

We argued that the CLEA loss helps learn more semantically meaningful representations. To illustrate the embeddings CLEA learns compared to the self-supervised approaches, we present examples from the visual modality in \autoref{fig:qualitative}. We selected the most similar image based on the cosine similarity of the embeddings. We show that the embeddings learned by CLEA qualitatively show more semantic similarity than self-supervised models. These images were selected using the 8-dimensional embeddings for all models.

For the idle behavior, self-supervised approaches show similar structural composition to the image of a robot, but CLEA methods show other robots or show similar flashing motifs. For the searching behavior, self-supervised approaches show similar square compositions to the reference image, but CLEA methods maintain map-like images. For the has information signal, the autoencoder recovered a similar lightbulb, whereas CLEA embeddings show other information-related items like books, or a pin identifying where an object is. For the has item signal, self-supervised approaches are unrelated to the idea of possessing an item, whereas CLEA methods show containers.

\subsection{Robustness to Noise}

\begin{figure*}
    \centering
    \includegraphics[width=\textwidth]{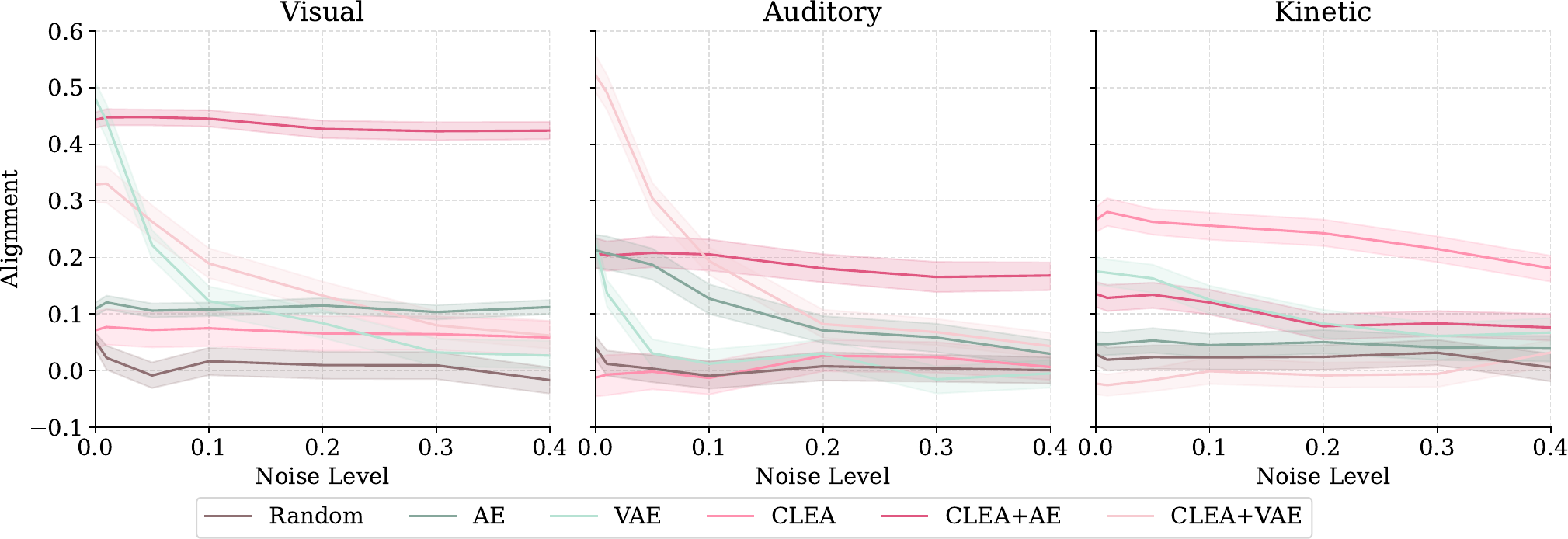}
    \caption{Robustness results. Accuracy of a linear reward model across different levels of injected noise. Features using CLEA maintain higher performance across different noise levels. We find that CLEA+AE is the least sensitive to noise overall.}
    \label{fig:noise}
\end{figure*}

Trajectories played on a real robot may have different actual features compared to the features initially associated with the trajectory due to differences in manufacturing or other physical effects.
To evaluate robustness to different forms of noise, we adopted a simple linear model, $R_H(\xi) = \omega \cdot \Phi(\xi)$ (the same model we used for evaluating simplicity and minimality in \autoref{sec:results}). We estimated $\omega$ using Bayesian inverse reward learning as in previous works \cite{biyik2019asking,sadigh2017active,brown2019extrapolating}. When learning from the observed user queries, we add noise from a uniform Gaussian to simulate inaccuracies in features for executed robot behaviors that slightly diverge from planned behaviors.

\begin{equation}
    \Phi(\xi)' = \Phi(\xi) + \epsilon \cdot \mathcal{N}(0,1)
\end{equation}

where $\epsilon$ represents a scaling factor of the noise. We evaluated final alignment after 100 queries using these modified features to assess robustness to noise. We used 8-dimensional features, modified the noise parameter with the values $\epsilon \in [0, 0.01, 0.05, 0.1, 0.2, 0.3]$, and averaged over 60 trials for each participants to control for random effects. The results are illustrated in \autoref{fig:noise}. 

In the visual modality, we found that CLEA+AE was the highest performing across all noise levels. In the auditory modality, CLEA+AE was the least sensitive to noise, but CLEA+VAE was more robust to noise level than a normal VAE, indicating that CLEA increases robustness. In the kinetic modality, CLEA was the highest performing across noise levels.

\subsection{Robustness to Sample Weighting}
\ndedit{Participants engaging in exploratory search may make more meaningful exploratory actions as they get closer to their desired signal. To evaluate if this is important in the learning process, we performed an additional experiment that weights exploratory actions at the end of the signal design process as more important than exploratory actions taken at the beginning of the signal design process. To weight the samples, we used the following equation based on the index $i$ of the sample ordered by time, and the total number of samples, $N$:}

\begin{equation}
    w(i) = \frac{i}{N}
\end{equation}

\ndedit{We trained each of the CLEA algorithms again, multiplying the loss for each datapoint by this additional weight value. We compared the weighted models---CLEA (weighted), CLEA+AE (weighted), CLEA+VAE (weighted)---with the unweighted models---CLEA, 
 CLEA+AE, CLEA+VAE---for each of the four criteria: completeness, minimality, simplicity, and explainability. We used the same evaluation processes as described in \autoref{sec:results}.}

\begin{table}[ht]
\centering
\caption{Comparison of Completeness, measured with test preference accuracy (TPA), for unweighted and weighted training. There were no significant differences in TPA for any modality or method (all p-values $> .05$).}
\label{tab:weighted_completeness}
\begin{tabular}{rr|ccc}
\midrule
 &  & TPA & TPA & \multicolumn{1}{l}{p-value} \\
 & & (unweighted)&(weighted)& (unc.)\\ \midrule
& CLEA & .955 & .949 & .553 \\
 Visual & CLEA+AE & .974 & .973 & .911 \\
 & CLEA+VAE & .810 & .841 & .383 \\
 \midrule
 & CLEA & .887 & .933 & .063 \\
 Auditory & CLEA+AE & .949 & .935 & .295 \\
 & CLEA+VAE & .800 & .767 & .429 \\
 \midrule
& CLEA & .976 & .973 & .789 \\
 Kinetic& CLEA+AE & .979 & .973 & .525 \\
 & CLEA+VAE & .936 & .934 & .904 \\\midrule
\end{tabular}
\end{table}

\ndedit{\textbf{Completeness.} To evaluate the effect of completeness, we used pairwise t-tests to assess differences in TPA between the weighted and unweighted models. The results are shown in \autoref{tab:weighted_completeness}. We observe no significant differences between any method or modality, indicating that there is no effect on feature completeness when reweighting samples.} 

\begin{table*}[t]
\centering
\caption{Comparison of AUC Alignment across unweighted and weighted variants of the CLEA algorithms. Numbers that perform better are in bold. We find that there were 22 of 45 trials where the unweighted CLEA performed the best, and 23 of 45 trials where the weighted CLEA performed the best. These values do not significantly differ from random chance ($p>.05$).}
\label{tab:minimality}

\resizebox{\textwidth}{!}{%

\begin{tabular}{rr|cc|cc|cc|cc|cc}
\midrule
 &  & {\scriptsize AUC Alignment} & {\scriptsize AUC Alignment}& {\scriptsize AUC Alignment}& {\scriptsize AUC Alignment}& {\scriptsize AUC Alignment}& {\scriptsize AUC Alignment}& {\scriptsize AUC Alignment}& {\scriptsize AUC Alignment}& {\scriptsize AUC Alignment}& {\scriptsize AUC Alignment} \\
 
 &  & (unweighted) & (weighted)  & (unweighted) & (weighted) & (unweighted) & (weighted) & (unweighted) & (weighted) & (unweighted) & (weighted) \\
 
 &  Dimension & \multicolumn{2}{c|}{8} & \multicolumn{2}{c|}{16} & \multicolumn{2}{c|}{\textbf{32}} & \multicolumn{2}{c|}{\textbf{64}} & \multicolumn{2}{c}{\textbf{128}} \\ \midrule
 & CLEA & -.011 & \textbf{.011} & .288 & \textbf{.386} & \textbf{.304} & .239 & \textbf{.187} & .116 & \textbf{.118} & .054 \\
 
 Visual & CLEA+AE & .440 & \textbf{.506} & .411 & \textbf{.459} & .380 & \textbf{.418} & .298 & \textbf{.333} & \textbf{.390} & .341 \\
 
 & CLEA+VAE & .257 & \textbf{.449} & .300 & \textbf{.422} & .379 & \textbf{.433} & \textbf{.317} & .222 & .156 & \textbf{.231} \\ \midrule
 
& CLEA & .075 & \textbf{.119} & -.078 & \textbf{.177} & \textbf{.166} & .063 & \textbf{.155} & -.027 & \textbf{.101} & -.071 \\

Auditory & CLEA+AE & .337 & \textbf{.373} & .005 & \textbf{.225} & .189 & \textbf{.275} & .018 & \textbf{.087} & .229 & \textbf{.253} \\

 & CLEA+VAE & \textbf{.524} & .279 & \textbf{.347} & .273 & \textbf{.303} & .156 & \textbf{.206} & .196 & \textbf{.176} & .141 \\ \midrule
 
& CLEA & \textbf{.289} & .265 & .329 & \textbf{.350}  & .295 & \textbf{.310} & \textbf{.416} & .403 & \textbf{.405} & .335 \\

 Kinetic & CLEA+AE & .092 & \textbf{.266} & .209 & \textbf{.288} & \textbf{.309} & .272 & \textbf{.396} & .351 & \textbf{.335} & .320 \\
 & CLEA+VAE & .000 & \textbf{.117} & \textbf{.265} & .178 & .161 & \textbf{.244} & \textbf{.400} & .379 & \textbf{.423} & .276\\
 \midrule
\end{tabular}
}
\end{table*}

\ndedit{\textbf{Minimality and Simplicity.} To evaluate the effect of minimality and simplicity, we present the results of AUC Alignment after 100 simulated pairwise queries. All values are significant because we can re-run simulations as many times as necessary, so we evaluate differences between the two algorithms by counting how many times the weighted version outperforms the weighted version for each trial. We use a binomial test to determine if the count of trial wins is significantly different from random chance, i.e., 50\%. For \textbf{simplicity}, we examined AUC Alignment across all dimensions, and found that unweighted CLEA models outperform weighted CLEA models in 22 of 45 trials, which is not significantly different than random chance ($p=.500$). For \textbf{minimality}, we examine just the smallest feature space dimension. We find that unweighted CLEA outperforms weighted CLEA in 2 of 9 trials, which is not significantly different than random chance, though it does trend toward significance ($p=.0912$), which warrants future exploration.}

\begin{table}[ht!]
\centering
\caption{Explainability }
\label{tab:explainability}
\begin{tabular}{rr|ccc}
\midrule
 &  & Similarity & Similarity & p-value \\
 &  & (unweighted)& (weighted) & (unc.) \\ \midrule
& CLEA & \textbf{.239} & .225 & \textbf{.003*} \\
 Visual & CLEA+AE & \textbf{.169} & .140 & \textbf{.001*} \\
 & CLEA+VAE & .136 & \textbf{.141} & .486 \\ \midrule
& CLEA & .248 & \textbf{.286} & \textbf{.001*} \\
 Auditory & CLEA+AE & .250 & \textbf{.253} & .835 \\
 & CLEA+VAE & .193 & \textbf{.259} & \textbf{.001*} \\ \midrule
& CLEA & .203 & \textbf{.204} & .896 \\
 Kinetic & CLEA+AE & .198 & \textbf{.202} & .213 \\
 & CLEA+VAE & \textbf{.307} & .281 & .188 \\ \midrule
\end{tabular}
\end{table}

\ndedit{\textbf{Explainability.} To evaluate the effect of weighted training on explainability, we calculated the cosine similarity of the top-ranked signals from the ranking study (\autoref{sec:eval_study}) to their nearest exemplar from the customization session (\autoref{sec:data_collection}) for both the weighted and unweighted CLEA variants. We present the results in \autoref{tab:explainability}. We found that there were only four significant differences across the three modalities. In two of the four differences, the unweighted version of CLEA showed a higher similarity, but this does not differ from random chance according to a binomial test ($p=.500$). Notably, the two instances where the unweighted CLEA training performed best were both in the Visual modality, and the two instances where weighted CLEA training performed best were in the Auditory modality. While we cannot draw any strong conclusions, these results may indicate that users explore different modalities in different ways. Future research may investigate reweighting strategies that capture these differences.}

\ndedit{\textbf{Summary.} Across the four evaluation criteria, we found that there was no clear benefit for using unweighted or weighted sampling techniques. Notably, CLEA works well without additional engineered training techniques such as sample re-weighting. This result underscores the simplicity of CLEA algorithms to leverage information from users' exploratory behaviors without having to explicitly model the user's search process.}

\subsection{Comparison with Learning Rewards without Features}

\begin{figure}
    \centering
    \includegraphics[width=\linewidth]{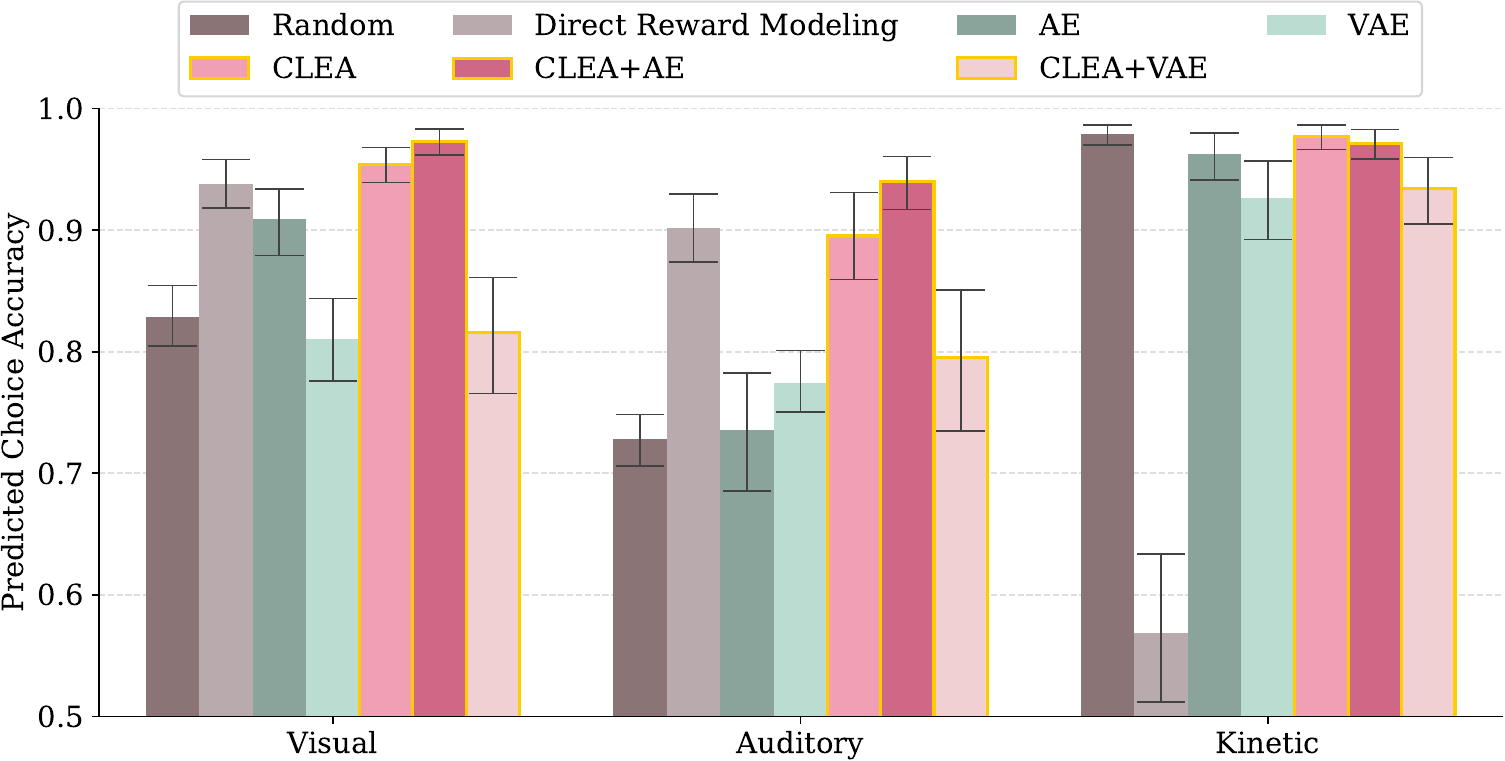}
    \caption{Direct reward modeling results. We show the completeness across the different methods of learning rewards, including directly learning rewards. We find that CLEA+AE outperforms all self-supervised methods and direct reward modeling.}
    \label{fig:direct_reward_modeling}
\end{figure}

To validate that learning features is useful, we tested learning a user's reward function from raw inputs. While this is not scalable as each user is required to first perform ten ranking tasks, this approach is highly expressive because it uses a large network that take the raw data structures as input. We used the same reward network as detailed in Appendix \ref{sec:reward_training}. The only change we made was to update the feature's model's weights (detailed in Appendix C) during training. We started with a feature learning network with randomly initialized weights, compared to the rest of our evaluations, where the weights of the feature network were frozen. We call this method ``Direct Reward Learning" and show the results in \autoref{fig:direct_reward_modeling}

Direct reward modeling showed improvement over self-supervised approaches, except for the kinetic modality.  This modality is highly prone to over-fitting, and with the small individual preference datasets, direct reward modeling is not able to learn a generalizable reward function for the user. In the visual modality, CLEA+AE achieved a mean accuracy of .973, compared to .938 for Direct Reward Modeling. In the Auditory modality, CLEA+AE achieved .940 compared to .902 for Direct Reward Modeling. In the Kinetic modality, CLEA+AE achieved .971 compared to .569 for Direct Reward Modeling. This result underscores the utility of using human-generated data to learn features to both facilitate downstream preference learning and more easily scale to large numbers of users.

\end{document}